\documentclass{article}
\pdfoutput=1
\usepackage[final]{neurips_2024}

\usepackage[utf8]{inputenc} 
\usepackage[T1]{fontenc}    
\usepackage{hyperref}       
\usepackage{url}            
\usepackage{booktabs}       
\usepackage{amsfonts}       
\usepackage{nicefrac}       
\usepackage{microtype}      
\usepackage{xcolor}         
\usepackage{graphicx}
\usepackage{amsmath}
\usepackage{multirow}
\usepackage{wrapfig}
\usepackage{siunitx}
\usepackage{colortbl}
\usepackage{hyperref}

\definecolor{deepPink}{RGB}{255, 20, 147}

\title{Grasp as You Say: \\Language-guided Dexterous Grasp Generation}

\author{
    Yi-Lin Wei\textsuperscript{1}, 
    \quad Jian-Jian Jiang\textsuperscript{1},
    \quad Chengyi Xing\textsuperscript{2},
    \quad Xian-Tuo Tan\textsuperscript{1},
    \\ \textbf{Xiao-Ming Wu\textsuperscript{1},}
    \quad \textbf{Hao Li\textsuperscript{2},}
    \quad \textbf{Mark Cutkosky\textsuperscript{2},}
    \quad \textbf{Wei-Shi Zheng\textsuperscript{1,3}\footnotemark[2]} \\ 
    \textsuperscript{1} School of Computer Science and Engineering, Sun Yat-sen University, China \\
    \textsuperscript{2} Stanford University, USA \\
    \textsuperscript{3} Key Laboratory of Machine Intelligence and Advanced Computing, 
    Ministry of Education, China \\ 
    \tt{
    \{weiylin5, jiangjj35, tanxt23, wuxm65\}@mail2.sysu.edu.cn 
    } \\
    \tt{\{chengyix, li2053, cutkosky\}@stanford.edu \quad wszheng@ieee.org}
}

\begin{document}
\maketitle
\footnotetext[2]{Corresponding author}
\vspace{-10mm}
\begin{center}
\href{https://isee-laboratory.github.io/DexGYS/}{\textcolor{deepPink}{https://isee-laboratory.github.io/DexGYS/}}
\end{center}
\vspace{8mm}
\begin{abstract}
This paper explores a novel task \textbf{``Dexterous Grasp as You Say''} (DexGYS), enabling robots to perform dexterous grasping based on human commands expressed in natural language. However, the development of this field is hindered by the lack of datasets with natural human guidance; thus, we propose a language-guided dexterous grasp dataset, named \textbf{DexGYSNet}, offering high-quality dexterous grasp annotations along with flexible and fine-grained human language guidance. Our dataset construction is cost-efficient, with the carefully-design hand-object interaction retargeting strategy, and the LLM-assisted language guidance annotation system. Equipped with this dataset, we introduce the \textbf{DexGYSGrasp} framework for generating dexterous grasps based on human language instructions, with the capability of producing grasps that are intent-aligned, high quality and diversity. To achieve this capability, our framework decomposes the complex learning process into two manageable progressive objectives and introduce two components to realize them. The first component learns the grasp distribution focusing on intention alignment and generation diversity. And the second component refines the grasp quality while maintaining intention consistency. Extensive experiments are conducted on DexGYSNet and real world environments for validation.
\end{abstract}

\section{Introduction}
Enabling robots to perform dexterous grasping based on human language instructions is essential within the robotics and deep learning communities, offering promising applications in industrial production and domestic collaboration scenarios.

With the advancements in data-driven deep learning and the availability of large-scale datasets, robot dexterous grasp methods achieve impressive performance~\cite{ddg,gendexgrasp,unigrasp,scene_diffuser,lu2023ugg,weng2024SingerViewDex,xu2024dgtr}. While previous approaches focus on the grasp stability, they have not fully utilized the potential of dexterous hands for intentional, human-like grasping. Recent studies, known as task-oriented~\cite{chen2023TaskDex} and functional dexterous grasping~\cite{zhu2023FunctionDex2, wei2023FunctionDex1}, aim to generate grasps based on specific tasks or functionality of objects. However, these approaches often depend on predefined, fixed and limited tasks or functions, restricting their flexibility and hindering natural human-robot interaction.

\begin{figure}[t]
\centering
\label{fig: new_task}
\includegraphics[width=\linewidth]{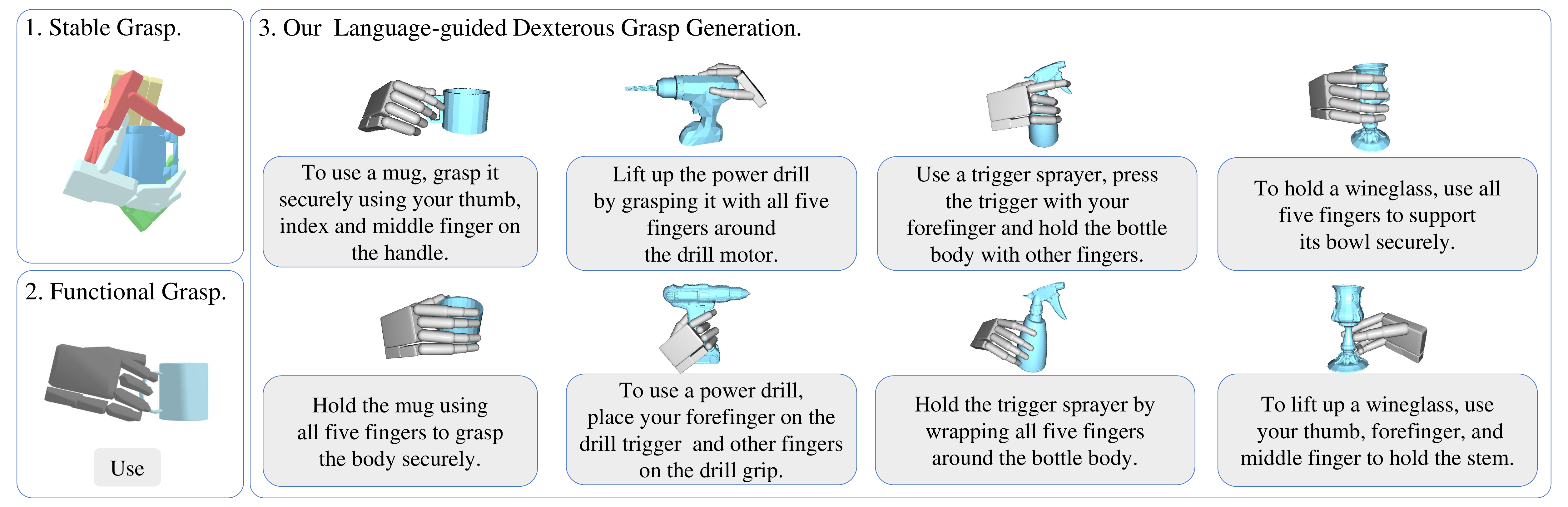}
\caption{Our Language-guided Task vs. Traditional Dexterous Grasp Tasks. Traditional methods focus either solely on grasp quality or on fixed and limited functionalities. Our approach enables the generation of dexterous grasps based on human language, enhancing natural human-robot interactions.
}
\end{figure}

In this paper, we explore a novel task, \textbf{``Dexterous Grasp as You Say''} (DexGYS), as shown in Figure \ref{fig: new_task}. We can see that natural human guidance is provided in this task, and can be utilized to drive dexterous grasping generation, thereby facilitating more user-friendly human-robot interactions. However, the new task also brings in new challenges. First, the high costs of annotating dexterous pose and the corresponding language guidance, present a barrier for developing and scaling dexterous datasets. Second, the demands of generating dexterous grasps that ensure intention alignment, high quality and diversity, present considerable challenges to the model learning.

To address the first challenge, we propose a large-scale language-guided dexterous grasping dataset \textbf{DexGYSNet}. DexGYSNet is constructed in a cost-effective manner by exploiting human grasp behavior and the extensive capabilities of Large Language Models (LLM). Specially, we introduce the Hand-Object Interaction Retargeting (HOIR) strategy to transfer easily-obtained human hand-object interactions to robotic dexterous hand, to maintain contact consistency and high-quality grasp posture. Subsequently, we develop the LLM-assisted Language Guidance Annotation system to produce flexible and fine-grained language guidance for dexterous grasp data with the support of LLM. DexGYSNet dataset comprises 50,000 pairs of high-quality dexterous grasps and their corresponding language guidance, on 1,800 common household objects.

With the support of the dataset, we now turn our way to overcome the second challenge. We propose the \textbf{DexGYSGrasp} framework for dexterous grasp generation, which aligns with intentions, ensures high quality, and maintains diversity. At the beginning, we find the difficulty of mastering all objectives simultaneously results from the commonly used penetration loss~\cite{xu2024dgtr} which used to avoid hand-object penetration. As shown in Figure \ref{fig: motivation1}, penetration loss substantially hinders the learning of grasp distribution, causing intention misalignment and reduced diversity. Conversely, despite the high diversity and aligned intention, the removal of penetration loss leads to unacceptable object penetration, making the grasp infeasible. Based on this finding, we design our DexGYSGrasp framework in a progressive strategy, decomposing the complex learning task into two sequential objectives managed by progressive components. Initially, the first component learns a grasp distribution, which focuses on intention consistency and diversity, optimizing effectively without the constraints of penetration loss. Subsequently, the second component refines the initial coarse grasps to high-quality ones with the same intentions and diversity. Our framework allows each component to focus on specific and manageable optimization objective, enhancing the overall performance of the generated grasps.

Extensive experiments are conducted on the DexGYSNet dataset and real-world scenarios. The results demonstrate that our methods are capable of generating intention-consistent, high diversity and high quality grasp poses for a wide range of objects.

\section{Related work}
\subsection{Dexterous Grasp Generation}
Dexterous hand endows robots with the capability to manipulate objects in a human-like manner. Previous methods have achieved impressive results in ensuring grasp stability by 
analytical approaches~\cite{analytic_1, DFC, q1, dexgraspnet, graspd} and deep learning methods~\cite{lu2023ugg, unigrasp, xu2024dgtr, unidexgrasp, efficientgrasp, generating_multi_finger}. However, the full potential of dexterous hands for intentional and human-like grasping has not been completely exploited in these methods. Recently, some works have focused on functional dexterous grasping~\cite{chen2023TaskDex, zhu2023FunctionDex2, wei2023FunctionDex1, zhu2021toward}, aiming to achieve human-like capabilities that extend beyond grasp stability alone, but are still lack of flexibility and generalization. In this work, we explore a novel task, Language-guided Dexterous Grasp Generation, which fully leverages the dexterity of robotic hands and enable robot to execute dexterous grasp based on human natural language.
\subsection{Grasp Datasets}
The development of large-scale datasets has significantly improve the advancement of data-driven grasp methods, including parallel grasp \cite{fang20201billion, eppner2021acronym, economic, tang2023task2_2, cai2024real}, human grasp \cite{hasson2019obman, chao2021dexycb, yang2022oakink, jian2023affordpose, li2024semgrasp, wang2024s2p}, and dexterous grasp approaches \cite{ddg, gendexgrasp, chen2023TaskDex, dexgraspnet, unidexgrasp}. Despite these advancements, the high cost of data collection remains a significant challenge, particularly in the domain of dexterous hands. Previous datasets for dexterous grasping primarily rely on physical analysis approaches \cite{DFC, miller2004graspit} to mitigate this issue. However, these approaches often lack the specific semantic context or corresponding language guidance necessary for constructing our language-guided dexterous task. In this paper, we present DexGYSNet dataset, with a cost-effective construction, providing high-quality dexterous grasp annotation along with flexible and fine-grained human language guidance.

\subsection{Language-guided Robot Grasp}
Language-guided robot grasp is important in robotics. Previous works focusing on parallel grippers have made strides in achieving task-oriented grasping~\cite{murali2021task2_1, tang2023task2_2, tang2023graspgpt}, language-guided grasping~\cite{xu2023joint, jin2024reasoning} and manipulation~\cite{jang2022bc, mees2022matters, driess2023palme, shridhar2021cliport}. In contrast to parallel grippers, dexterous hand boast a higher number of DOF (e.g., 28 for the Shadow Hand~\cite{shadowhand}), enabling a broader dexterity. However, this high freedom also presents challenges for model learning. In this paper, we propose the DexGYSGrasp framework, capable of generating intention-aligned dexterous grasps with high-quality and diversity.

\begin{figure}[t]
\centering
\includegraphics[width=\linewidth]{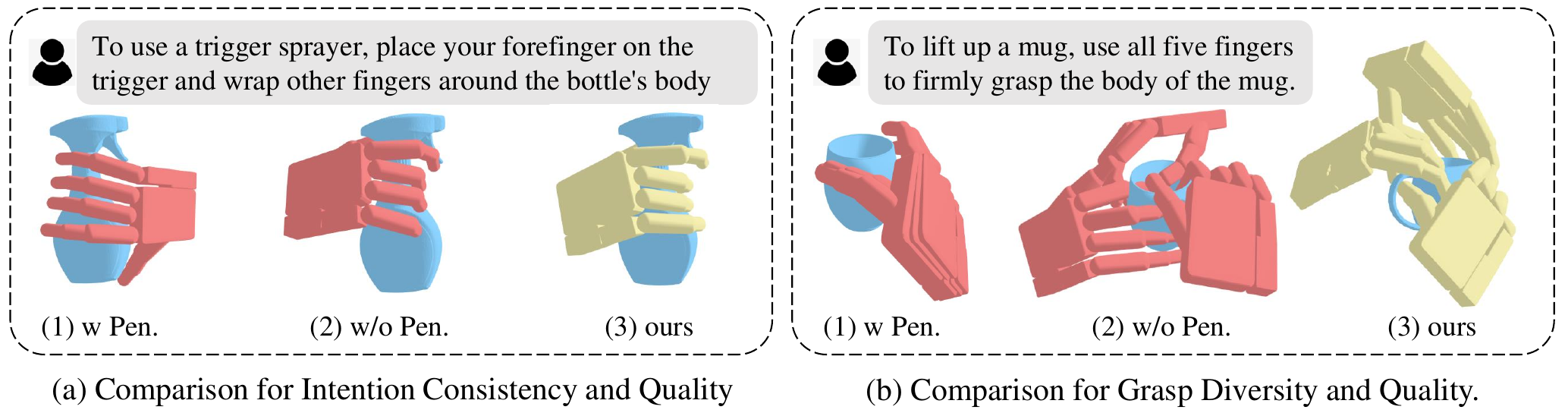}
\vspace{-5mm}
\caption{
Visualization of the impact of penetration loss (Pen. in the figure) on grasp performance: intention alignment, quality, and diversity. (a) illustrates penetration loss \textbf{causes intention misalignment} and its absence results in severe object penetration. (b) shows three sampling results under the same conditions, and demonstrates that penetration loss \textbf{leads to reduced diversity}.
}
\label{fig: motivation1}
\end{figure}

\section{DexGYSNet Dataset}
\subsection{Dataset Overview}

\begin{figure}[t]
\centering
\includegraphics[width=\linewidth]{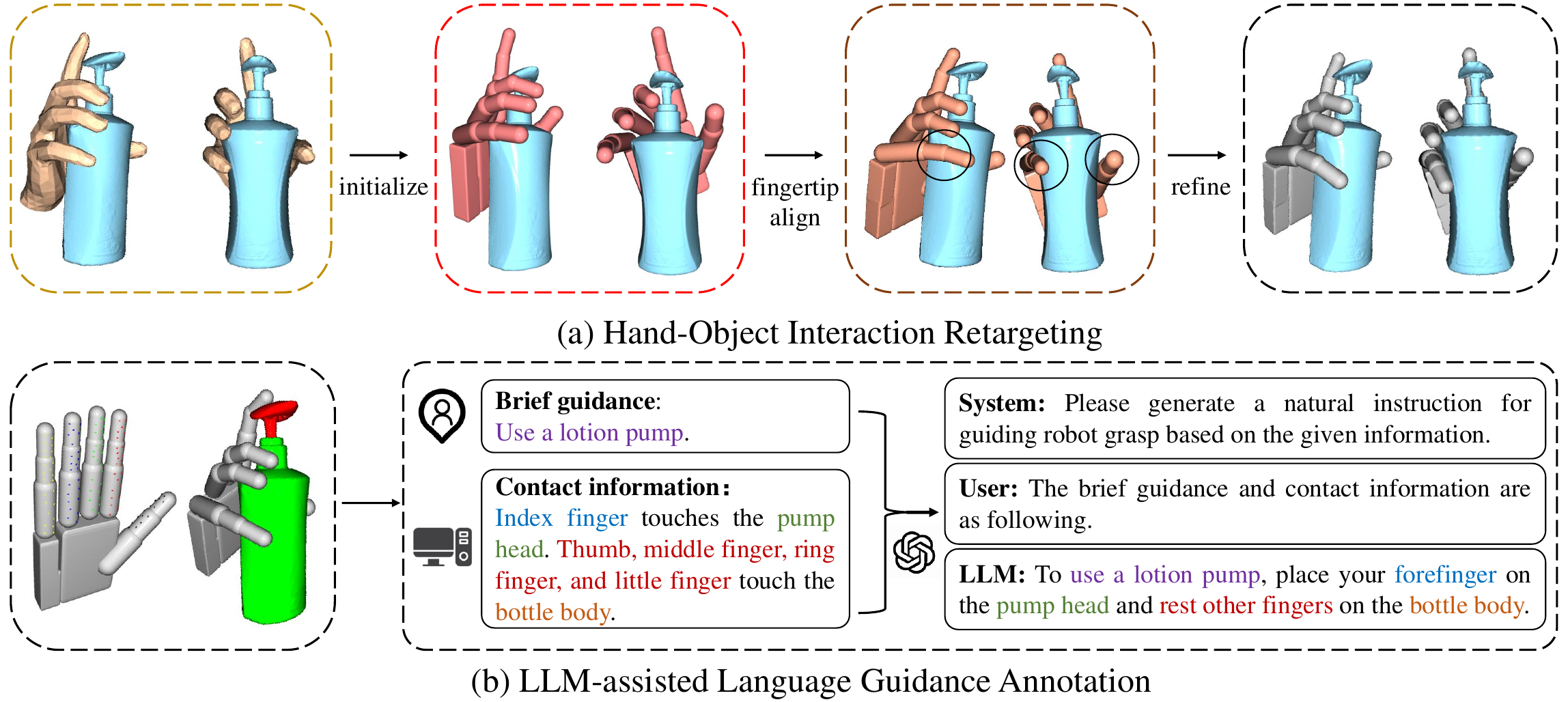}
\vspace{-4mm}
\caption{
The construction process of the DexGYSNet dataset. (a) The HOIR strategy retargets the human hand to the dexterous hand by three step, maintaining hand-object interaction consistency and avoiding physical infeasibility (shown in black circle). (b) The annotation system automatically annotates language guidance for hand-object pairs with the help of LLM.
}
\label{fig: dexdysconstruction}
\end{figure}

The DexGYSNet dataset is constructed with a cost-effective strategy, as shown in Figure \ref{fig: dexdysconstruction}. We first collect object meshes and human grasps data from existing datasets~\cite{yang2022oakink}. Subsequently, we develop the Hand-Object Interaction Retargeting (HOIR) strategy to transform human grasps into dexterous grasps with high quality and hand-object interaction consistency. Finally, we implement an LLM-assisted Language Guidance Annotation system, which leverages the knowledge of Large Language Models (LLM) to produce flexible and fine-grained annotations for language guidance.


\subsection{Hand-Object Interaction Retargeting}
\label{section:HOIR}
Our Hand-Object Interaction Retargeting (HOIR) aims to transfer human hand-object interaction to dexterous hand-object interaction as shown in Figure~\ref{fig: dexdysconstruction} . The source MANO \cite{romero2017MANO} hand parameters are denoted as $\mathcal{G}^{m}\in {R}^{61}$. And the target dexterous hand parameters are denoted as $\mathcal{G}^{dex}$ = $(r, t, q)$,  where $r \in \mathbf{SO(3)}$ represents the global rotation, $t \in \mathbb{R}^{3}$ is the translation in world coordinates, and $q \in \mathbb{R}^{J}$ is the joint angles for a $J$-DoF dexterous hand, for example $J = 22$ for Shadow Hand\cite{shadowhand}.

Three steps are within the HOIR: pose initialization, fingertip alignment, and interaction refinement. In the first step, the dexterous poses are initialized by copying parameters from similar structures of human poses to establish better initial values. In the second step, the dexterous poses are optimized in the parameter space to align the fingertip positions $p_{k}^{dex,ft}$ with those of the human $p_{k}^{mano,ft}$. This achieves retargeting consistency, and the optimization objective can be formulated as follows:
\begin{equation}
\label{eq:step2}
\setlength{\abovedisplayskip}{5pt}
\setlength{\belowdisplayskip}{5pt}
\min_{\mathcal{G}^{dex}=(r,t,q)} {\sum_{k}{\|p_{k}^{dex,ft}-p_{k}^{mano,ft}\|_2^2}}.
\end{equation}
To improve the physical interaction feasibility while maintaining the consistency, the dexterous hand poses are further optimized in the third step by hand-object interaction and physical constraints losses~\cite{dexgraspnet}. Two key points are designed to maintain the consistency: preserving the contact area of the optimized pose consistent with the output from the second step, and keeping the translation fixed during this step. The optimization objective can be formulated as follows:
\begin{equation}
\label{eq:step3}
\setlength{\abovedisplayskip}{5pt}
\setlength{\belowdisplayskip}{3pt}
\min_{(r,q)}{(\lambda^{1}_{pen}\mathcal{L}_{pen}+\lambda^{1}_{spen}\mathcal{L}_{spen}+\lambda^{1}_{joint}\mathcal{L}_{joint}+\lambda^{1}_{cmap}\mathcal{L}_{cmap})}.
\end{equation}
Here, the object penetration loss $\mathcal{L}_{pen}$ penalizes the depth of hand-object penetration. The self-penetration loss $\mathcal{L}_{spen}$ penalizes the self-penetration. The joint angle loss $\mathcal{L}_{joint}$ penalizes the out-of-limit joint angles. The contact map loss $\mathcal{L}_{cmap}$ ensures the contact map on the object remains consistent with the output from the second stage. The details of losses can be found in Appendix~\ref{sec: loss details}.

\subsection{LLM-assisted Language Guidance Annotation}
To annotate flexible and fine-grained language guidance for dexterous hand-object pairs with low-cost, we design a coarse-to-fine automated language guidance annotation system with the assistance of the LLM, inspired by~\cite{cui2024anyskill,li2024semgrasp}, as shown in Figure~\ref{fig: dexdysconstruction}. Specially, we initially generate brief guidance based on the object category and the brief human intention (e.g., "using a lotion pump"), which are collected by the human dataset~\cite{yang2022oakink}. Subsequently, we compile the contact information for each finger by calculating the distances from the contact anchors on the hand to different parts of the object. We then organize the contact information into language descriptors (e.g. "forefinger touches pump head and other fingers touch the bottle body."). Finally, we input both the brief guidance and the detailed contact information into the GPT3.5 to produce natural annotated guidance (e.g. "To use a lotion pump, press down on the pump head with your forefinger while holding the bottle with your other fingers."). More details about DexGYSNet construction can be found in Appendix~\ref{sec: dexgys details}.

\begin{figure}[t]
\centering
\includegraphics[width=\linewidth]{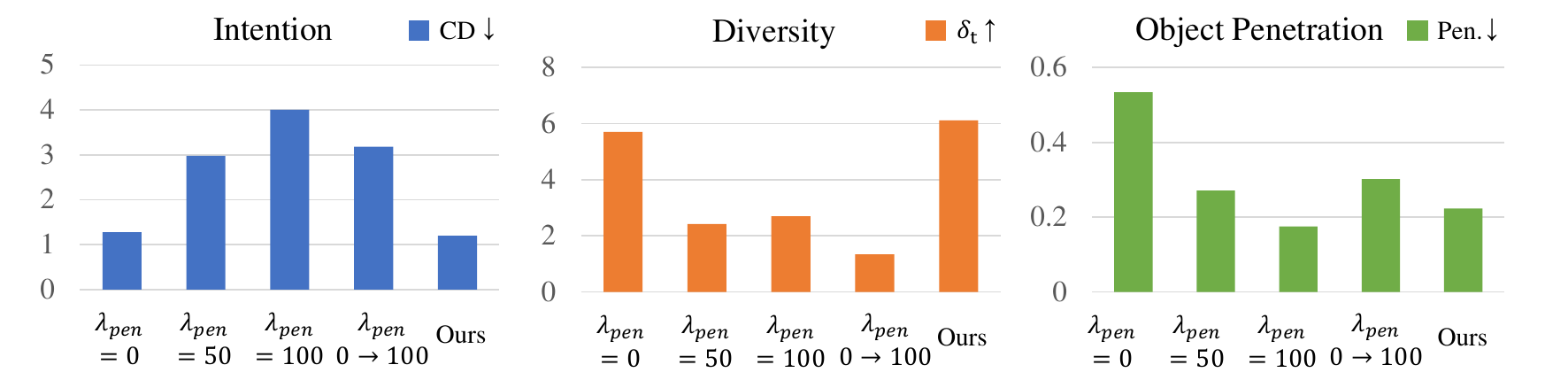}
\caption{
Quantitative experimental results with different object penetration loss weights $\lambda_{pen}$. Intention is quantified by the Chamfer distance (CD) between predictions and targets. Diversity is assessed by the standard deviation of hand translation $\delta_{t}$. Object penetration is evaluated by the penetration depth (Pen.) from the object point cloud to the hand mesh. Our method uniquely achieves high performance in terms of intention consistency, diversity, and penetration avoidance.
}
\label{fig: motivation2}
\end{figure}

\section{DexGYSGrasp framework}
Given full object point clouds $\mathcal{O}$ and language guidance $\mathcal{L}$ as inputs, our goal is to generate dexterous grasps $\mathcal{G}^{dex}$ with intention alignment, high diversity and high quality. 

\subsection{Progressive Grasp Objectives.}
\label{sec: learning challenge}
\textbf{Learning Challenge in DexGYS.} The DexGYS places high demands on intention alignment (e.g., accurately pressing your forefinger on trigger to use the sprayer), high diversity (e.g., holding the bottle using various postures), and high quality (e.g., ensuring stable grasp and avoiding object penetration). However, we find that a single model struggles to meet these requirements simultaneously, due to the optimization challenge caused by the commonly used object penetration loss~\cite{grasptta, dexgraspnet, unidexgrasp}, which is used to prevent hand-object penetration. As shown in Figure~\ref{fig: motivation1} and Figure~\ref{fig: motivation2}, increasing the weight of the penetration loss reduces object penetration but adversely affects intention alignment and generation diversity. 

\textbf{Progressive Grasp Objectives.} To address these challenges, we propose to decompose the complex learning objective into two more manageable objectives. The first objective is generative: it focuses on learning the grasp distribution, which does not prioritize quality but focuses on learning the grasp distribution with intention alignment and generation diversity. The second objective is regressive: it aims to refine the coarse grasp to a specific high-quality grasp with same intention. By decomposing the complex objectives, we reduce the learning difficulty of the generative objective as it does not concentrate on quality and avoids using penetration loss which could interfere the learning process. Additionally, the learning of regression is less complex than distributions, as it merely requires adjusting the pose to a specific target within a small space. Hence, we can employ penetration loss to ensure that the refined dexterous hand avoids penetrating the object and with high quality. 

\subsection{Progressive Grasp Components}
Benefiting from our progressive grasp objectives in Section~\ref{sec: learning challenge}, we design the following two simple progressive grasp components, which can achieve intention alignment, high diversity and high quality language-guided dexterous generation.

\begin{figure}[t]
\centering
\includegraphics[width=\linewidth]{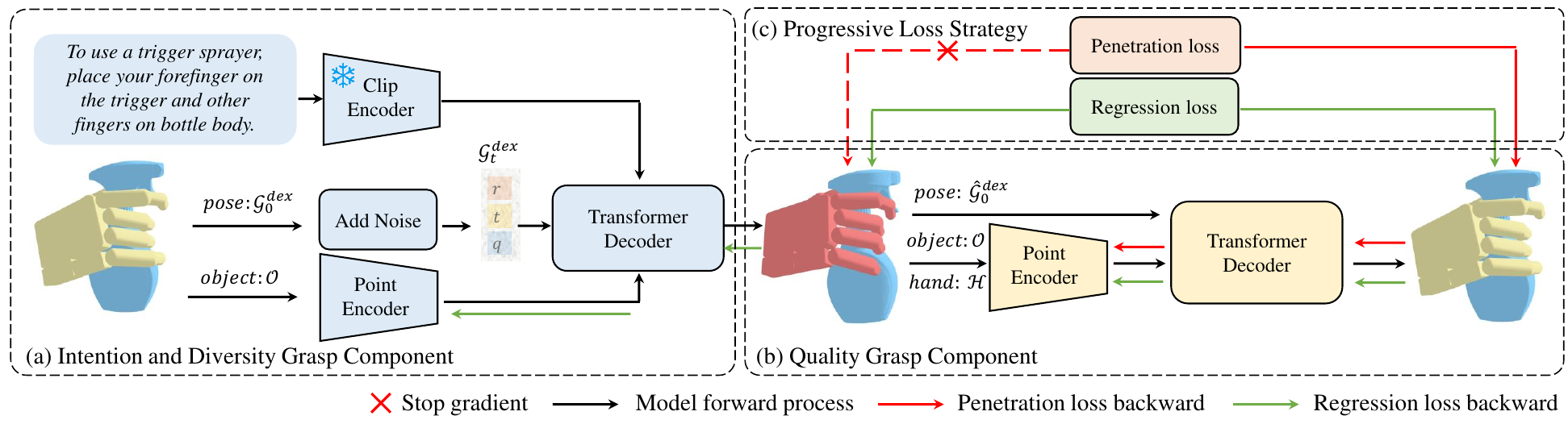}
\caption{
Overview of our framework. (a) With only the regression loss, intention and diversity grasp component is trained to reconstruct the original hand pose from the noise poses, based on language and object condition. (b) With both regression and penetration losses, Quality Grasp Component is trained to refine the coarse pose improve the grasp quality while maintain intension consistency.
}

\label{fig: methods}
\end{figure}
\textbf{Intention and Diversity Grasp Component.} We introduce intention and diversity grasp component to learn a grasp distribution efficiently, achieve intention aligned and diverse generation. Due to the distribution modeling objective, IDGC is build upon the conditional diffusion model \cite{condition_diffusion_1, scene_diffuser} to predict the dexterous pose $\mathcal{G}^{dex}_{0}$ from noised $\mathcal{G}^{dex}_{T}$. The input object point clouds $\mathcal{O}$ is encoded by Pointnet++~\cite{pointnet++} and language $\mathcal{L}$ is encoded by a pretrained CLIP model \cite{radford2021clip} as the condition. And we employ DDPM \cite{ho2020DDPM} as sampling process, which can be formalized by the following equation:
\begin{equation}
p_{\theta}\left(\mathcal{G}^{dex}_{0} | \mathcal{O}, \mathcal{L}\right) = p\left(\mathcal{G}^{dex}_{T}\right) \prod_{t=1}^T p\left(\mathcal{G}^{dex}_{t-1}|\mathcal{G}^{dex}_{t}, \mathcal{O}, \mathcal{L}\right).
\end{equation}

\textbf{Quality Grasp Component.} The generated grasps of the first component possess well-aligned intentions and high diversity, but suffer from poor grasp quality due to significant object penetration. Therefore, we introduce Quality Grasp Component to refine the grasp quality while maintaining intention consistency in a regressive manner. Specially, it takes the coarse pose $\hat{\mathcal{G}}^{dex}$, coarse hand point clouds $\mathcal{H}(\hat{\mathcal{G}}^{dex})$ and object point clouds $\mathcal{O}$ as input, and outputs the pose $\Delta\mathcal{G}^{dex}$. The refined grasp is obtained by $\tilde{\mathcal{G}}^{dex}=\hat{\mathcal{G}}^{dex}+\Delta\mathcal{G}^{dex}$. The training pairs of this component are constructed by collecting coarse grasps generated by the first component alongside the most similar ground-truth grasps that share the similar intentions. This ensures the training targets are aligned with the language intention, thereby guaranteeing that the refined grasps maintain consistency with the intended actions.  

\subsection{Progressive Grasp Loss}
\textbf{Intention and Diversity Grasp Loss.} We strategically employ regression losses and exclude object penetration loss to enhance the training efficacy of intention and diversity grasp component. By focusing exclusively on the regression learning, this component facilitates a more effective optimization process, achieving enhancements of intention consistency and grasp diversity. Concretely, we utilize L2 loss for pose parameter regression and incorporate the hand chamfer loss \cite{chamfer} to assist by explicit hand shape. The loss function of intention and diversity grasp component.is defined as:
\begin{equation}
\begin{split}
&\mathcal{L}_{IDG} = \lambda_{para}^{2}\mathcal{L}_{para}(\mathcal{G}^{dex}_{0}, \hat{\mathcal{G}}^{dex}) + \lambda_{chamfer}^{2}\mathcal{L}_{chamfer}(\mathcal{H}(\mathcal{G}^{dex}_{0}), \mathcal{H}(\hat{\mathcal{G}^{dex}})), \\
\end{split}
\end{equation}
where $\mathcal{H}$ are dexterous hand point clouds of corresponding pose. 

\textbf{Quality Grasp Loss.} Benefiting from the simplified training objectives, the quality grasp component focuses solely on refining coarse grasp to a specific target within a relatively constrained space, thereby reducing the negative impact of object penetration. Therefore, we employ the well-designed loss including object penetration. The loss function of quality grasp component can be formulated as:
\begin{equation}
\label{eq:IRT}
\mathcal{L}_{QG} = \lambda^{3}_{para}\mathcal{L}_{para}+\lambda^{3}_{chamfer}\mathcal{L}_{chamfer}+\lambda^{3}_{pen}\mathcal{L}_{pen}+\lambda^{3}_{cmap}\mathcal{L}_{cmap}+\lambda^{3}_{spen}\mathcal{L}_{spen}.
\end{equation}
More details about loss function and model structure can be found in Appendix~\ref{sec: model details}.

\section{Experiments}
\subsection{Datasets and Evaluation Metrics}
We split the DexDYSNet dataset at the object instance level, using 80\% of the objects within each category for training and 20\% for evaluation. Notably, none of the objects in the test set appear in the training set, ensuring that all experimental results are evaluated on unseen objects.

Three types of metrics are employed for evaluation from the perspective of intention consistency, grasp quality and grasp diversity. 1) For intention consistency, we employ \textbf{Fréchet Inception Distance} (FID), using sampling point cloud features extracted from~\cite{nichol2022pointe}  to calculate $P\text{-}FID$ and rendering image features extracted from~\cite{heusel2017fid} to calculate $FID$. Additionally, \textbf{Chamfer distance} ($CD$), is used to measure the distance between predicted hand point clouds and targets; \textbf{Contact distance} ($Con.$) is used to measure the L2 distance of object contact map between the prediction and targets. 2) For grasp quality, \textbf{Success rate} in Issac gym and $\mathbf{Q_{1}}$~\cite{q1} measure grasp stability. We set the contact threshold to \SI{1}{cm} and set the penetration threshold to \SI{5}{mm} following~\cite{dexgraspnet}. \textbf{Maximal penetration depth} (\text{cm}), denoted as $Pen.$, reflects the maximal penetration depth from the object point cloud to hand meshes. 3) For diversity, we employ the \textbf{Standard deviation} of translation $\delta_t$, rotation $\delta_r$ and joint angle $\delta_q$ of eight samples within same condition, following~\cite{xu2024dgtr}. More details can be found in Appendix~\ref{sec: Metrics Detials}.

\begin{table*}[t]
    \footnotesize
    \setlength{\tabcolsep}{3pt}
    \centering
        \begin{tabular}{c|*{4}{c}|*{3}{c}|*{3}{c}}
        \toprule
        \multirow{2}*{Method} & \multicolumn{4}{c|}{Intention} & \multicolumn{3}{c|}{Quality} & \multicolumn{3}{c}{Diversity} \\
        & $FID \downarrow$ & $P\text{-}FID \downarrow$ & $CD \downarrow$ & $Con. \downarrow$ & $Success \uparrow$ & $Q_{1}\uparrow$ & $Pen. \downarrow$ & $\delta_{t}\uparrow$ & $\delta_{r}\uparrow$ & $\delta_{q}\uparrow$ \\
        \midrule
        GraspCVAE\cite{cvae} &31.26 & 29.02 &3.138 &0.096 & 29.12\% & 0.054 &0.551 & 0.179  & 1.762  &0.179    \\
        GraspTTA\cite{grasptta}  & 35.41 & 33.15 &12.19 &0.111  & 43.46\% &0.071 & 0.188 &2.111  &6.150  &3.869   \\
        SceneDiffuser\cite{scene_diffuser} &20.44 &7.932 &1.679 &0.045 &62.24\% &0.083 & 0.253 &0.346 &3.455  &0.387   \\
        DGTR\cite{xu2024dgtr} &23.31 &15.77 &2.895 &0.078 & 51.91\% &0.078 &\textbf{0.163} &2.037  &14.01  &4.299   \\    
        \midrule
        Ours & \textbf{6.538} & \textbf{5.595} &\textbf{1.198} & \textbf{0.036} &\textbf{63.31\%} &\textbf{0.083} &0.223 &\textbf{6.118}  & \textbf{55.68} & \textbf{6.118} \\
        \bottomrule
        \end{tabular}
    \caption{Results on DexGYSNet compared with the SOTA methods.}
    \label{table:sota}
\end{table*}

\subsection{Implementation Details}
For the construction of DexGYSNet, the step 2 and 3 are optimized for 20 and 300 iterations with learning rates of 0.01 and 0.0001 respectively. We set $\lambda^{1}_{pen}=100$ and set $\lambda^{1}_{spen}$, $\lambda^{1}_{joint}$, $\lambda^{1}_{cmap}$ each to 10. For training our framework, the training epochs are set to 100 for intention and diversity grasp component and 20 for Quality Grasp Component. The loss weights are configured as follows: $\lambda^{2}_{para}=\lambda^{3}_{para}=10$, $\lambda^{2}_{chamfer}=\lambda^{3}_{chamfer}=1$, $\lambda^{3}_{cmap}=10$, $\lambda^{3}_{pen}=100$, $\lambda^{3}_{spen}=10$. Throughout all training processes, the model is optimized with a batch size of 64 using the Adam optimizer, with a weight decay rate of $5.0 \times 10^{-6}$. The initial learning rate is $2.0 \times 10^{-4}$ and decay to $2.0 \times 10^{-5}$ using a cosine learning rate~\cite{sgdr} scheduler. All experiment are implemented with PyTorch on a single RTX 4090 GPU. 

\begin{figure}[t]
\centering
\includegraphics[width=\linewidth]{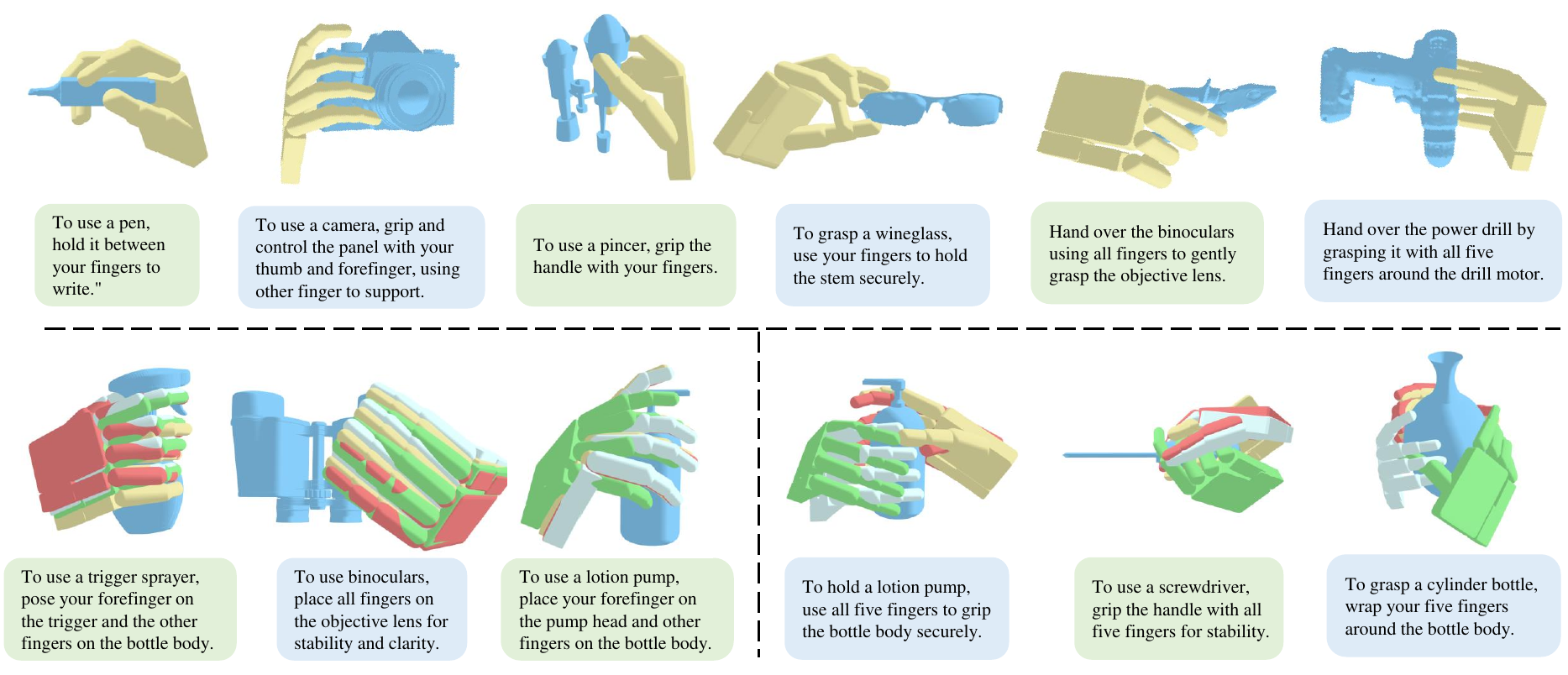}
\vspace{-3mm}
\caption{
Visualization of generated dexterous grasp. The \textbf{top} visualizes one sample for each object and guidance pair. The bottom visualizes four samples, the \textbf{bottom left} shows that the generated grasp are consistent with clear and specific guidance, while the \textbf{bottom right} shows that the diversity achieved under relatively ambiguous instructions.}
\label{fig:vis}
\end{figure}

\subsection{Comparison with SOTA methods}
The comparison results are presented in Table \ref{table:sota}. We reproduce the SOTA methods to suit our task by concatenating the language condition with the point cloud features, the details can be found in Appendix~\ref{sec: Implementation Details of SOTA Methods}. As seen in the Table, our framework significantly outperforms all previous methods in terms of intention consistency and grasp diversity,  while also achieving comparable performance in grasp quality. Previous methods struggle with learning a robust language conditional grasp distribution due to the optimization challenges outlined in Section \ref{sec: learning challenge}. They often yield misaligned yet high quality grasps, resulting in comparable grasp quality, but less aligned intention and limited diversity compared to our framework. Overall, these results confirm that our framework achieves SOTA performance in generating intention-aligned, high-quality and diverse grasps.

In Figure \ref{fig:vis}, we visualize the generated grasp to qualitatively demonstrate the grasp generation capabilities of our framework. The bottom figure visualizes the results of four samples, the bottom left highlights our framework's ability to produce precise and consistent grasps under deterministic guidance (e.g., the way to use a trigger sprayer is deterministic). In the other hand, the bottom right illustrates our framework's diversity in generating grasps when provided with ambiguous guidance (e.g., the way to hold a bottle is diverse). 

\begin{table*}[t]
    \footnotesize
    \setlength{\tabcolsep}{6.58pt}
    \centering
        \begin{tabular}{c|*{2}{c}|*{2}{c}|*{3}{c}}
        \toprule
        \multirow{2}*{} & \multicolumn{2}{c|}{Intention} & \multicolumn{2}{c|}{Quality} & \multicolumn{3}{c}{Diversity} \\
        & $CD \downarrow$ & $Con. \downarrow$ & $Q_{1}\uparrow$ & $Pen. \downarrow$ & $\delta_{t}\uparrow$ & $\delta_{r}\uparrow$ & $\delta_{q}\uparrow$ \\
        \midrule
        IDGC ($\lambda_{pen}^2=0$) &1.276 &0.028 & 0.024 &0.534 &5.710  &54.75  & 7.741   \\
        IDGC ($\lambda_{pen}^2=50$)  &2.980 &0.061 &0.074 &0.271 &2.421  &33.27  &3.391  \\
        IDGC ($\lambda_{pen}^2=100$) &4.009 &0.067 &0.072 &0.175   &2.701  &38.27  &3.785  \\
        IDGC ($\lambda_{pen}^2=500$) &4.185 &0.072 &0.107 &0.037 &0.547  &8.807  &0.481   \\
        \midrule
        IDGC $(\lambda_{pen}^2=0\rightarrow100$) &3.181 &0.056 &0.093  &0.302  &1.341 &16.53 &2.211     \\
        IDGC ($\lambda_{pen}^2=0$) + TTA &20.09 &0.102  & 0.057 &0.178  &4.849  &51.91  &8.479  \\
        IDGC ($\lambda_{pen}^2=100$) + QGC &2.009 &0.042 &0.099 &0.143 &3.414  &40.35  &2.844      \\
        \rowcolor{gray!25} 
        IDGC ($\lambda_{pen}^2=0$) + QGC &1.198 & 0.036 & 0.083 &0.223  &6.118  & 55.68 & 6.118 \\
        \bottomrule
        \end{tabular}
    \caption{Ablation study for our framework. Intention and diversity grasp component is abbreviated as IDGC, Quality Grasp Component is abbreviated as QGC. $\lambda^{2}_{pen}$ is the penetration loss weight tn the training of IDGC. Ours is colored in \textcolor{gray}{gray}.}
    \label{table:AB}
\end{table*}

\subsection{Necessity of Progressive Components and Losses}
The results presented in Table \ref{table:AB} validate the core insight of our framework: decomposing the complex task into progressive objectives, employing progressive components, and learning with progressive losses. The initial four lines of results demonstrate that a single component, without progressive objectives, fails to balance all objectives. Moreover, a single component, even with progressive objectives, that adjusts $\lambda_{pen}^2$ from $0$ to $100$ after several training epochs, does not enhance performance. The similar result occurs when using progressive components without corresponding progressive losses, $IDGC (\lambda_{pen}^2=100) + QGC$. Moreover, the commonly used quality refinement strategy test-time adaptation (TTA)~\cite{grasptta}, though improves grasp quality but results in extremely poor intention consistency. Overall, only the progressive designs of our DexGYSGrasp framework ensures excellence in intention alignment, high quality and diversity.

\begin{table*}[t]
    \parbox{.43\textwidth}{
        \centering
        \footnotesize
        \setlength{\tabcolsep}{3pt}
        \begin{tabular}{c c c | c | c c}
            \toprule
            \multicolumn{3}{c|}{} & Intention & \multicolumn{2}{c}{Quality} \\ 
            step1 & step2 & step3 & $Con.\downarrow$ & $Q_{1}\uparrow$ & $Pen \downarrow$ \\
            \midrule
            $\checkmark$ & & & 0.048 & 0.037 & 0.572 \\
            & $\checkmark$ & & 0.101 & 0.833 & 0.516 \\
            $\checkmark$ & $\checkmark$ & & 0.012 & 0.029 & 0.477 \\
            \rowcolor{gray!25} 
            $\checkmark$ & $\checkmark$ & $\checkmark$ & 0.015 & 0.063 & 0.369 \\
            \midrule
            \multicolumn{3}{c|}{\textit{all in one stage}} & 0.075 & 0.090 & 0.271 \\
            \multicolumn{3}{c|}{\textit{w/o fix translation}} & 0.051 & 0.074 & 0.332 \\
            \bottomrule
        \end{tabular}
        \caption{Ablation study for HOIR.}
        \label{table:HOIR}
    }
    \hfill
    \parbox{.55\textwidth}{
        \centering
        \footnotesize
        \setlength{\tabcolsep}{3pt}
        \begin{tabular}{p{1.8cm} | c c | c c | c}
            \toprule
            & \multicolumn{2}{c|}{Intention} & \multicolumn{2}{c|}{Quality} & Diversity \\ 
            & $CD\downarrow$ & $Con.\downarrow$ & $Q_{1}\uparrow$ & $Pen \downarrow$ & $\delta_{t}\uparrow$ \\
            \midrule 
            GraspCVAE & 3.138 & 0.096 & 0.054 & 0.551 & 0.179 \\
            + w/o $\mathcal{L}_{pen}$ & 2.638 & 0.090 & 0.005 & 0.921 & 0.266 \\
            + QGC & 2.425 & 0.056 & 0.074 & 0.261 & 0.315 \\
            \midrule
            SceneDiffuser & 1.679 & 0.045 & 0.083 & 0.225 & 0.346 \\
            + w/o $\mathcal{L}_{pen}$ & 1.511 & 0.041 & 0.019 & 0.488 & 6.323 \\
            + QGC & 1.495 & 0.040 & 0.082 & 0.241 & 6.015 \\
            \bottomrule
        \end{tabular}
        \caption{Plug-and-play Experiments.}
        \label{table:PPE}
    }
    \hfill
\end{table*}

\subsection{Plug-and-play Experiments}
We conducted experiments to evaluate the applicability of our insights to other state-of-the-art (SOTA) methods. Specifically, we trained GraspCAVE and SceneDiffuser without the object penetration constraint and trained the quality grasp component (QGC) to refine the coarse outcomes. As depicted in Table \ref{table:PPE}, removing the object penetration loss leads to improved intention consistency, which corroborates our findings discussed in Section \ref{sec: learning challenge}. Moreover, our quality grasp component can significantly enhance grasp quality while maintaining the intention consistency.

\subsection{Effectiveness of Hand-Object Interaction Retargeting}
We conducted ablation studies to evaluate our Hand-Object Interaction Retargeting (HOIR) strategy in constructing DexGYSNet dataset. As shown in Table \ref{table:HOIR}, our three-step HOIR significantly improves both the quality and the intention consistency progressively. We observed that optimizing all losses in Equations \ref{eq:step2} and \ref{eq:step3} in one step (\textit{all in one stage}), results in worse contact consistency and better grasp quality. Similar outcomes occur when the root translation is not fixed in step 3 (\textit{w/o fix translation}). We believe this trade-off arises from inherent noise in the hand-object interaction data and the structural differences between human grasps and dexterous hands, making it challenging to excel in all aspects. Overall, we think that three-step HOIR strategy achieves more comprehensive outcomes, especially in the most important aspect of hand object contact consistency.

\begin{figure}[t]
\centering
\includegraphics[width=\linewidth]{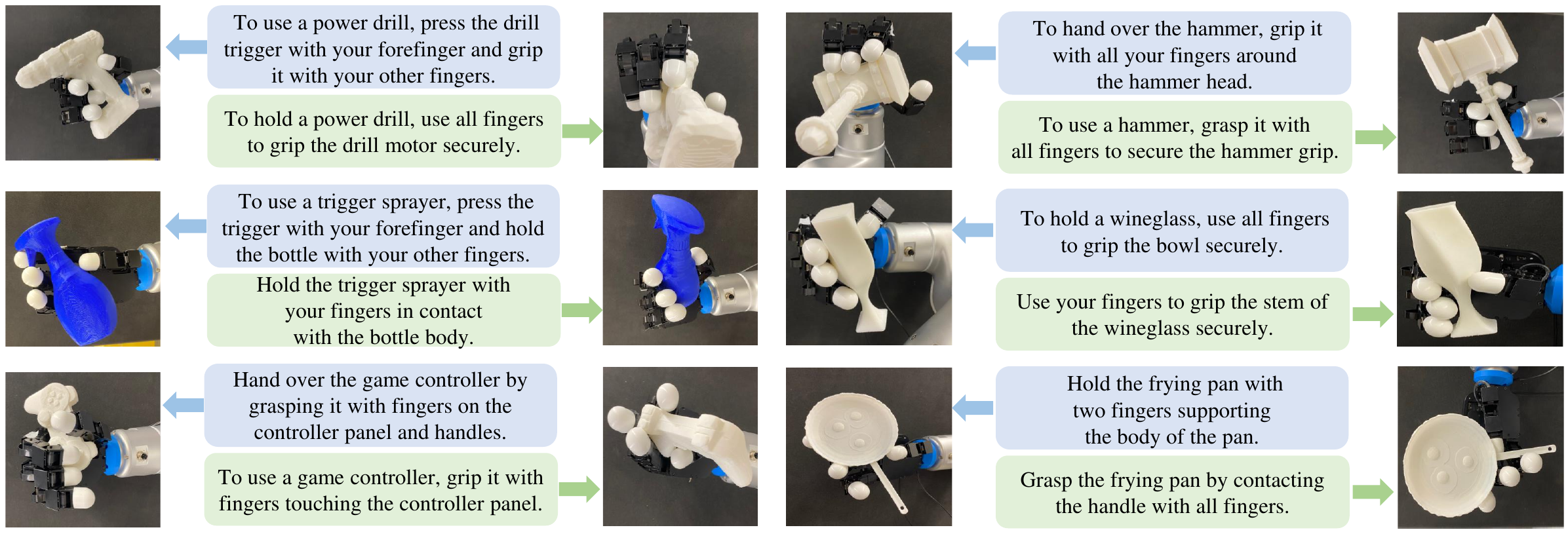}
\caption{
Visualization of real world experiments.
}
\label{fig: real world}
\end{figure}

\subsection{Experiments in Real World}
We conducted real-world grasp experiments to verify the practical application of our methods, as shown in Figure~\ref{fig: real world}. The experiments are conducted on an Allegro hand, a Flexiv Rizon 4 arm and an Intel Realsense D415 camera. Although our framework is designed for full object point clouds, we integrate several off-the-shelf methods to enhance its practicality. Specifically, partial object point clouds are obtained through visual grounding~\cite{liu2023groundingdino} and SAM~\cite{kirillov2023SAM}, which are then fed into a point cloud completion network~\cite{yuan2018pcn} to obtain full point clouds. In execution, we first move the arm to the 6-DOF pose of the dexterous hand root node, and then control the dexterous hand joint angles to the predicted poses. Real world experiments further validate the effectiveness of our method. More implementation details can be found in Appendix~\ref{sec: realworld detials}. 

\section{Conclusions}
We believe that enabling robots to perform high quality dexterous grasps aligned with human language is crucial within the deep learning and robotics communities. 
In this paper, we explore this novel task, ``Dexterous Grasp as You Say'' (DexGYS). This task is non-trival, we propose a DexGYSNet dataset and a DexGYSGrasp framework to accomplish it. DexGYSNet dataset is constructed cost-effectively using the object-hand interaction retargeting strategy and the language guidance annotation system assisted by LLMs. Building on DexGYSNet, DexGYSGrasp framework, comprised of two progressive components, which can achieve intention-aligned, high diversity, and high quality dexterous grasp generation. Extensive experiments in DexGYSNet and real-world settings demonstrate that our framework significantly outperforms all SOTA methods, confirming the potential and effectiveness of our approach.

\section*{Acknowledgements}
This work was supported partially by NSFC(92470202, U21A20471), Guangdong NSF Project (No. 2023B1515040025). Additionally, I sincerely thank the help of Guo-Hao Xu and Dian Zheng for the valuable suggestions for the paper.

{
\small
\bibliography{neurips_2024.bib}
\bibliographystyle{unsrt}
}


\newpage
\appendix

\section{Appendix / supplemental material}

\subsection{DexGYSGrasp Details}
\label{sec: model details}
\subsubsection{Diffusion Background}
The diffusion model is used in our intention and diversity grasp component to generate grasp distribution with aligned intention and high diversity, which represents a class of generative models characterized by a forward process of noise addition and a reverse process of denoising. The forward process entails a Markov Chain that incrementally introduces Gaussian noise into the data across multiple time steps. Originating from the initial data $x_0$, this process transitions the data to conform with a standard Gaussian distribution $x_T$ after $T$ time steps. This transformation is mathematically formulated as follows:
\begin{equation}
    x_t = \sqrt{\alpha_t} x_{t-1} + \sqrt{1-\alpha_t} \epsilon_{t-1}, 
\end{equation}
where $\alpha_t$ denotes a time-dependent noise coefficient, $\bar{\alpha}_t = \prod_{i=1}^t \alpha_i$.

Therefore, $x_t$ given $x_0$ follows a normal distribution,
\begin{equation}
q(x_{t}|x_{0})=\mathcal{N}(x_{t};\sqrt{\bar{\alpha}_{t}}x_{0},(1-\bar{\alpha}_{t})\mathbf{I}).
\end{equation}
The first equation delineates the stepwise diffusion, whereas the second equation offers a direct approximation of any intermediate state $x_t$ from $x_0$.

During the reverse process, the model is trained to closely approximate the reverse conditional distribution $p(x_{t-1} | x_t)$, which is described as:
\begin{equation}
p_\theta(x_{t-1} | x_t) =
\mathcal{N}(x_{t-1}; \mu_\theta(x_t, t), \sigma^2_\theta(x_t, t)),
\end{equation}
where $\mu_\theta(x_t, t)$ and $\sigma^2_\theta(x_t, t)$ are the mean and variance parameters for the distribution of $x_{t-1}$, respectively, and $\theta$ indicates the parameters of the model used to predict $\epsilon$ from $x_t$.

The classical sampling strategy for the reverse process is exemplified by DDPM \cite{ho2020DDPM}, where the model iteratively learns to reverse the noise addition process to reconstruct the original data from noise. It estimates the distribution $p(x_{t-1} | x_t)$ and predicts the noise $\epsilon$, represented by:
\begin{align}
    \boldsymbol{\mu}_\theta\left(\mathbf{x}_t, t\right) & = \tilde{\boldsymbol{\mu}}_t\left(\mathbf{x}_t, \frac{1}{\sqrt{\bar{\alpha}_t}}\left(\mathbf{x}_t - \sqrt{1-\bar{\alpha}_t} \boldsymbol{\epsilon}_\theta\left(\mathbf{x}_t\right)\right)\right) \\
    & = \frac{1}{\sqrt{\alpha_t}}\left(\mathbf{x}_t - \frac{\beta_t}{\sqrt{1-\bar{\alpha}_t}} \boldsymbol{\epsilon}_\theta\left(\mathbf{x}_t, t\right)\right), \\
    x_{t-1} & = \boldsymbol{\mu}_\theta\left(\mathbf{x}_t, t\right) + \sigma_\theta(x_t, t)z,
\end{align}
where $\sigma_\theta$ consists of non-trainable, time-dependent constants, and $z$ represents Gaussian noise.

\subsubsection{Intention and Diversity Grasp Component}
\textbf{Point Encoder}
We utilize a three-layer PointNet++~\cite{pointnet++} as our point encoder, following recent works~\cite{graspness, dexgraspnet, gendexgrasp, xu2024dgtr} in the field of robotic grasping. Specifically, each layer $l_i, i \in {1,2,3}$, receives point clouds and corresponding features (initially the raw XYZ coordinates for the first layer) from the preceding layer. It then performs down-sampling and feature aggregation using the "set-aggregation" operation~\cite{pointnet++}. The aggregated features are processed by a three-layer perceptron, which consists of three $Linear-BatchNorm-ReLU$ blocks. The output of point encoder is $\mathcal{F}_{obj} \in \mathbb{R}^{N_{obj}\times C_{obj}}$.

\textbf{Language Encoder}
For the language encoder, we employ the CLIP model with the ViT-L/14 architecture~\cite{radford2021clip}. The input text sequence is tokenized and converted into token embeddings with positional embeddings added. This sequence is processed through multiple Transformer encoder layers to obtain the language feature $\mathcal{F}_{lan} \in \mathbb{R}^{N_{lan}\times C_{lan}}$.

\textbf{Transformer Decoder}
We employ four layers of MLPs and Transformer as decoder\cite{transformer, scene_diffuser}. The time embedding and pose feature are incorporated in MLPs to obtain $\mathcal{F}_{dex\_t}$. Subsequently, the $\mathcal{F}_{dex\_t}$ serves as the query, and the concatenated features of language and object, $\mathcal{F}_{lan\_obj}$, act as key and value in Transformer block. The corss attention process is formalized as:
\begin{equation}
\mathcal{F}_{out} = \text{softmax}\left(\frac{f_{q}(\mathcal{F}_{dex\_t})f_{k}(\mathcal{F}_{lan\_obj})^{T}}{\sqrt{d_k}}\right)f_{v}(\mathcal{F}_{lan\_obj}),
\end{equation}
where ${f}_{q}, {f}_{k}, {f}_{v}$ are MLPs, and $d_k$ is the channel of features. Finally, a MLP is adopted to regress the dexterous grasp parameters $\hat{\mathcal{G}}^{dex}$. 
 $\mathcal{F}_{dex\_t} = {f}_{0}(f_{1}({F}_{dex})+f_{2}(\mathcal{F}_{time}))$, where ${f}_{0}, {f}_{1}, {f}_{2}$ are MLPs.

\subsubsection{Quality Grasp Component}
Quality grasp component tasks coarse pose $\hat{\mathcal{G}}^{dex}$, coarse hand point clouds $\mathcal{H}(\hat{\mathcal{G}}^{dex})$ and object point clouds $\mathcal{O}$ as input, and outputs refined grasp $\tilde{\mathcal{G}}^{dex}$. The object and hand are encoded by the PointNet++ same with intention and diversity grasp component. And in the transformer decoder, coarse pose features act as the query, object and hand features serve as key and value. 

\begin{figure}[t]
\centering
\includegraphics[width=\linewidth]{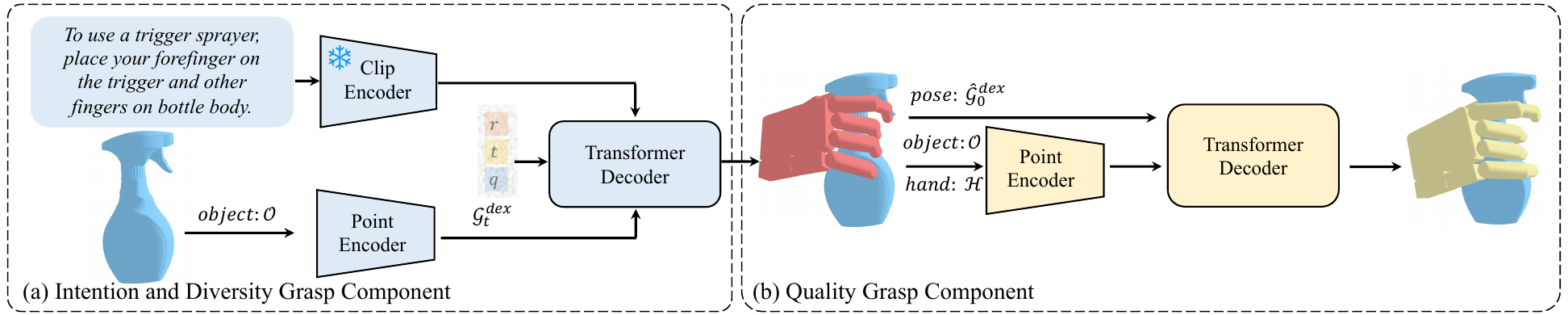}
\caption{
Inference pipeline of our DexGYSGrasp.
}
\label{fig:inference}
\end{figure}

\subsubsection{Inference Pipeline}
We also demonstrate the inference pipeline of our DexGYSGrasp, as shown in Fig. \ref{fig:inference}. We sample a random noise from Gaussian distribution as the input, with the point cloud and language guidance as the conditions. We first generate the coarse grasp by the intention and diversity grasp component, and then refine it with the quality grasp component.

\subsubsection{Loss Function}
\label{sec: loss details}
This section provides a detailed exposition of the loss functions utilized during the construction of datasets and the training of models.

\textbf{Parameter Regression Loss.}
We utilize the mean squared error (MSE) to quantify the deviation between the generated dexterous hand pose $\hat{\mathcal{G}}^{dex}$ and the ground truth $\mathcal{G}^{dex}$.
\begin{equation}
\mathcal{L}_{para} = \frac{1}{N} \sum_{i=1}^N \|\mathcal{G}^{dex,i} - \hat{\mathcal{G}^{dex,i}}\|_2^2.
\end{equation}

\textbf{Hand Chamfer Loss.}
The predicted hand point clouds $\mathcal{H}(\hat{\mathcal{G}}^{dex})$ and the ground truth $\mathcal{H}(\mathcal{G}^{dex})$ are derived by sampling from the hand mesh. We then compute the chamfer distance to assess the discrepancies between the predicted and ground-truth hand shapes.
\begin{equation}
\label{chamferloss}
    \mathcal{L}_{chamfer} 
    = \sum_{x \in \mathcal{H}(\mathcal{G}^{dex})} \min_{y \in \mathcal{H}(\hat{\mathcal{G}}^{dex})} \|x - y\|_{2}^{2}
    + \sum_{x \in \mathcal{H}(\hat{\mathcal{G}}^{dex})} \min_{y \in \mathcal{H}(\mathcal{G}^{dex})} \|x - y\|_{2}^{2}.
\end{equation}

\textbf{Contact Map Loss.}
The contact map loss $\mathcal{L}_{cmap}$ ensures consistency between the predicted hand contact map $\hat{c}^{obj}$ on object and the target $c^{obj}$. The contact map is calculate by the distance from object point to the closest dexterous hand point.
\begin{equation}
\label{cmaploss}
\mathcal{L}_{cmap} = \sum_{i} \|c_{i}^{obj} - \hat{c}_{i}^{obj}\|_2^2.
\end{equation}

\textbf{Object Penetration Loss.}
The object penetration loss $\mathcal{L}_{pen}$ penalizes the depth of hand-object penetration, where $d_{i}^{sdf}$ denotes the signed distance from the object point to the hand mesh.
\begin{equation}
\mathcal{L}_{pen} = \sum_{i} \mathbb{I}(d_{i}^{sdf} > 0) \cdot d_{i}^{sdf}.
\end{equation}

\textbf{Self-Penetration Loss.}
The self-penetration loss $\mathcal{L}_{spen}$ punishes the penetration among the different parts of the hand, where $p^{dex,sp}$ denotes predefined anchor spheres on the hand~\cite{dexgraspnet}.
\begin{equation}
\mathcal{L}_{spen} = \sum_{i,j} \mathbb{I}(i \neq j) \cdot \max(\delta - d(p_{i}^{dex,sp}, p_{j}^{dex,sp})).
\end{equation}

\textbf{Joint Angle Loss.}
Given the physical structure limitations of the robotic hand, each joint has designated upper and lower limits. The joint angle loss penalizes deviations from these limits.
\begin{equation}
\mathcal{L}_{joint} = \sum_{i} (\max(q_{i} - q_{i}^{max}) + \max(q_{i}^{min} - q_{i})).
\end{equation}

\begin{figure}[t]
\centering
\includegraphics[width=\linewidth]{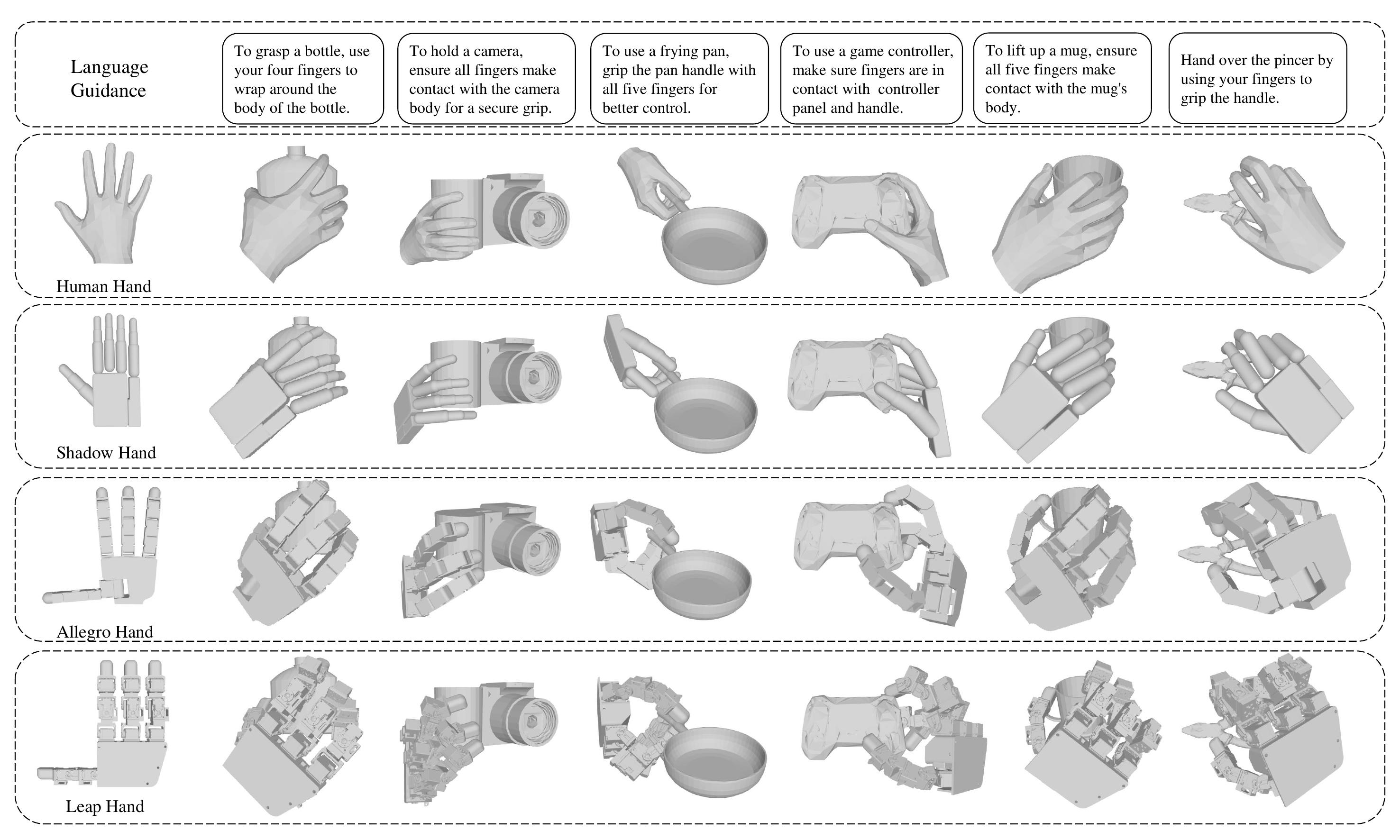}
\caption{
The Extension of DexGYSNet to more dexterous hands.
}
\label{fig: morehand}
\end{figure}

\subsection{DexGYSNet Datasets Details}
\label{sec: dexgys details}
\subsubsection{Prompt of LLM}
We introduce the prompt for using GPT-3.5 in this section.

\textbf{System Prompt}: "You are an assistant in creating language instruction, aimed at guiding robot on how to grasp objects. Given a brief instruction and a fine-gained interaction information. Your task is generate a natural and more informative instruction. The instruction should start with the given brief instruction, which is limited in a sentence and about 10-15 words."

\textbf{User Prompt}: "Brief instruction: To <\textit{brief intention}> a <\textit{object category}>. Hand-object interaction information: <\textit{contact information}>. " 

The \textit{brief intention} and \textit{object category} are sourced from the hand-object dataset OakInk~\cite{yang2022oakink}. The \textit{contact information} is derived by calculating the distances from predefined contact anchors on each finger to the segmentation parts of the object. Details on predefined contact anchors are available in DexGraspNet~\cite{dexgraspnet}, and segmentations are annotated in OakInk~\cite{yang2022oakink}.

An example of user prompt is: "Brief instruction: To use a trigger sprayer. Hand-object interaction information: forefinger touches the trigger. thumb, middle finger, ring finger and little finger touches the finger." An example of LLM output is: "To use a trigger sprayer, press the trigger with your forefinger and hold the bottle with your other fingers."

\subsubsection{DexGYSNet Extension}
Our cost-effective dataset construction strategy can be easily extended to various types of dexterous hands. As shown in Figure~\ref{fig: morehand}, besides the Shadow Hand\cite{shadowhand}, which features a highly biomimetic design replicating most degrees of freedom of human hands at a high cost of \$100,000, we also expand to the Allegro Hand~\cite{allegrohand} and Leap Hand~\cite{shaw2023leap}. These latter models, while offering fewer degrees of freedom, are significantly more affordable, costing \$16,000 and \$2,000, respectively, making them practical for promoting the use of robotic arms in laboratory environments. We have trained our method on the DexGYSNet dataset using the Allegro Hand and implemented it in real robot experiments.

\begin{figure}[t]
    \centering
    \includegraphics[width=\linewidth]{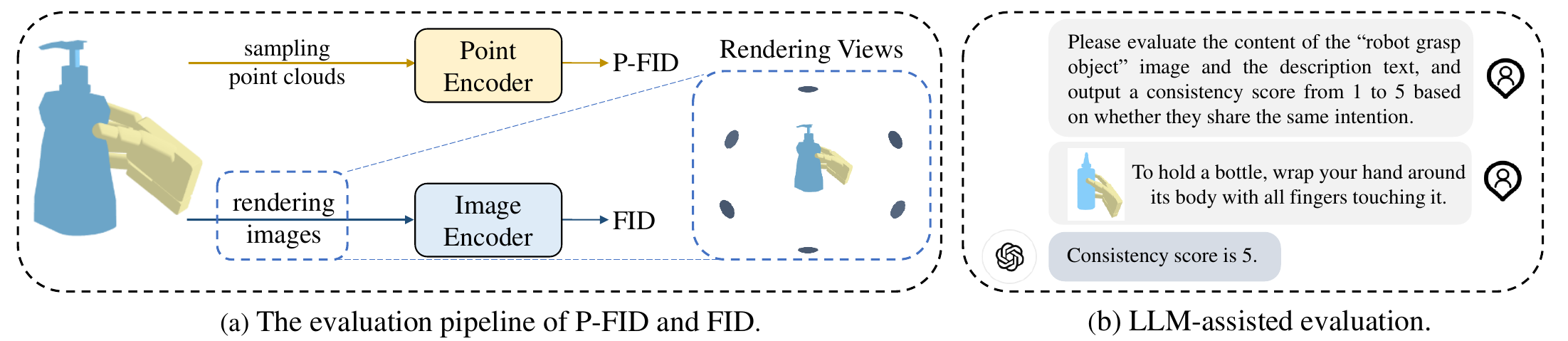}
    \vspace{-0.5cm}
    \caption{(a) Evaluation of intention consistency using the Fréchet Inception Distance between the \textless generation hand and object\textgreater\,and the ground truth. (b) When the ground truth is not available (e.g., evaluation on a 3D object dataset), we employ GPT4-o for evaluation.}
    \label{fig: fid}
\end{figure}

\subsection{Implementation Details}
\label{sec: Implementation Details}
\subsubsection{Dataset Split}
We split the DexDYS dataset at the level of object instances. Specially, for all objects within each category, 80\% of the objects instances are used for training and 20\% for evaluation. Concretely, the training set includes approximately 1,200 objects with 40k grasps, while the evaluation set comprises about 300 objects with 10k grasps. Therefore, all objects in the test set of DexGYSNet don't exist in the training set.

\subsubsection{Metrics Detials}
\label{sec: Metrics Detials}
\textbf{Target Assignment.} For target assignment in the testing phase, the grasp targets of an object-guidance pair consist of all poses that share the same contact part and brief guidance. And the matrices of intention consistency are calculated by comparing the prediction to the most similar grasp target.

\textbf{Fréchet Inception Distance}, which is commonly used in generative task\cite{guo20223Dhuman} by measuring the distance between the generated distribution and the ground truth distribution. We use sampling point cloud features extracted from\cite{nichol2022pointe} to calculate P-FID and rendering image features extracted from\cite{heusel2017fid} to calculate FID. The details are shown in Figure~\ref{fig: fid} (a).

\textbf{Chamfer distance,} denoted as $CD$, is used to measure the distance between predicted hand point clouds and targets to measure the consistency from the aspect of hand consistency. Please look at Equation~\ref{chamferloss} for details.

\textbf{Contact distance,} denoted as $Con.$ to measure the L2 distance of object contact map between the prediction and targets to measure the consistency from the aspect of object contact consistency. Please look at Equation~\ref{cmaploss} for details.

\textbf{Success rate.}
We evaluate the grasp success rate in Issac Gym simulation environment. To simulate the force exerted by dexterous hands grasping objects in real environments, we contract each finger in the direction of the object. If the grasp can withstand at least one of the six directions of gravity, it is considered successful.

\textbf{Mean $Q_{1}$.}
Intuitively, the $Q_{1}$ metric reflects the norm of the smallest wrench which can disrupt the stability of a grasp. We follow~\cite{dexgraspnet} to set the contact threshold to 1cm and set the penetration threshold to 5mm. Any grasp with its maximal penetration depth greater than 5mm is considered invalid and we set the $Q_{1}$ of it to 0 before taking the average.

\textbf{Maximal penetration depth,} which is the maximal penetration depth from the object point cloud to hand meshes.

\textbf{Diversity.} We use the standard deviation of translation $\delta_t$, rotation $\delta_r$, and joint angle $\delta_q$ to measure the diversity of generated grasps. We perform eight samples in the intention and diversity component under the same input conditions, and each sample is individually sent to the quality component for refinement. Before calculation, $\delta_r$ and $\delta_q$ are converted to Euler angles in degrees, while $\delta_t$ is measured in centimeters. 

\subsubsection{Implementation Details of SOTA Methods}
\label{sec: Implementation Details of SOTA Methods}
We replicate SOTA methods on our DexGYSNet dataset using the same encoder structure and the loss functions defined in Equation \ref{eq:IRT} to ensure fair comparison. Specifically, we reimplement GraspCAVE based on \cite{cvae}, GraspTTA \cite{grasptta}, SceneDiffuser \cite{scene_diffuser}, and DGTR \cite{xu2024dgtr}. To introduce language information, we use an identical CLIP language encoder. For GraspCAVE, we concatenate the language feature, object feature, and latent feature to send to the decoder. Based on GraspCAVE, GraspTTA employs a test-time adaptation strategy for quality refinement. For SceneDiffuser, we concatenate the language and object features as the model condition. For DGTR, the language and object features are concatenated to send to its transformer decoder.

\begin{figure}[t]
\centering
\includegraphics[width=0.9\linewidth]{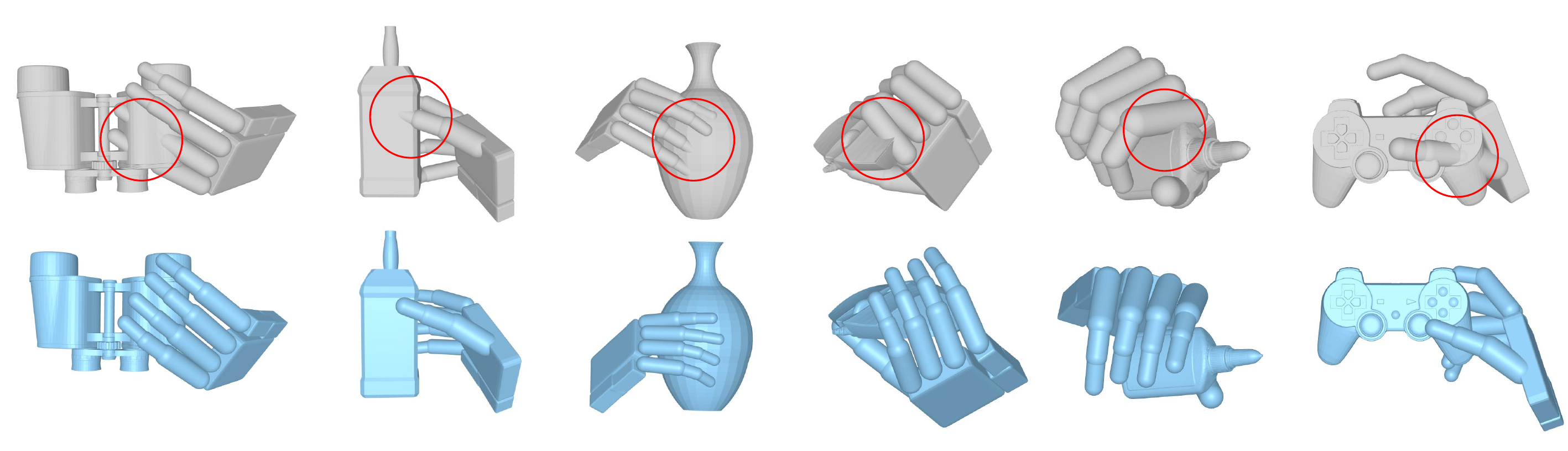}
\caption{
Visualization of grasps before and after quality grasp component. Our quality grasp component improves grasp quality and maintains intention consistency.
}
\label{fig: coarse2fine}
\end{figure}

\begin{figure}[t]
\centering
\includegraphics[width=0.9\linewidth]{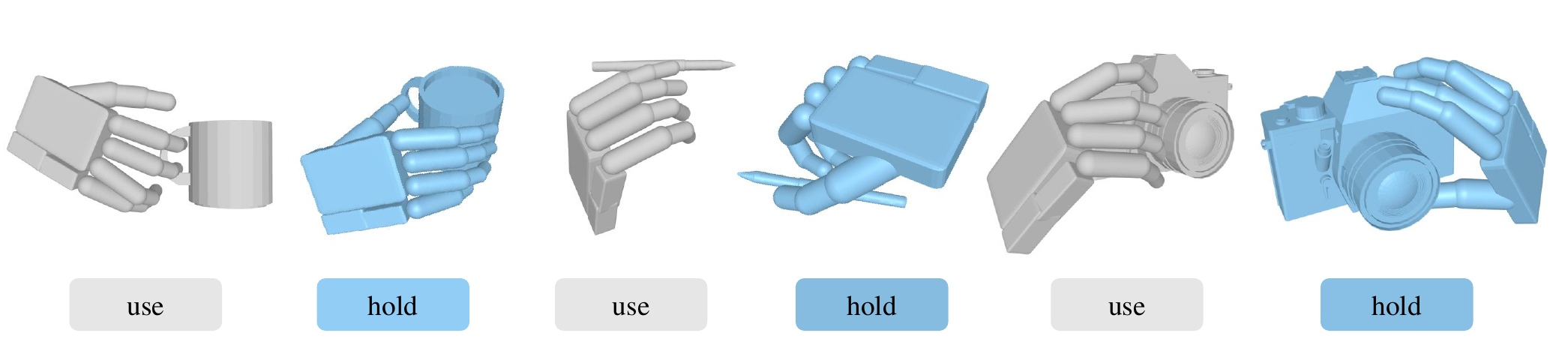}
\caption{
Visualization of our DexGYSGrasp framework with task-oriented simple input.
}
\label{fig: simple_input}
\end{figure}

\begin{figure}[t]
\centering
\includegraphics[width=0.8\linewidth]{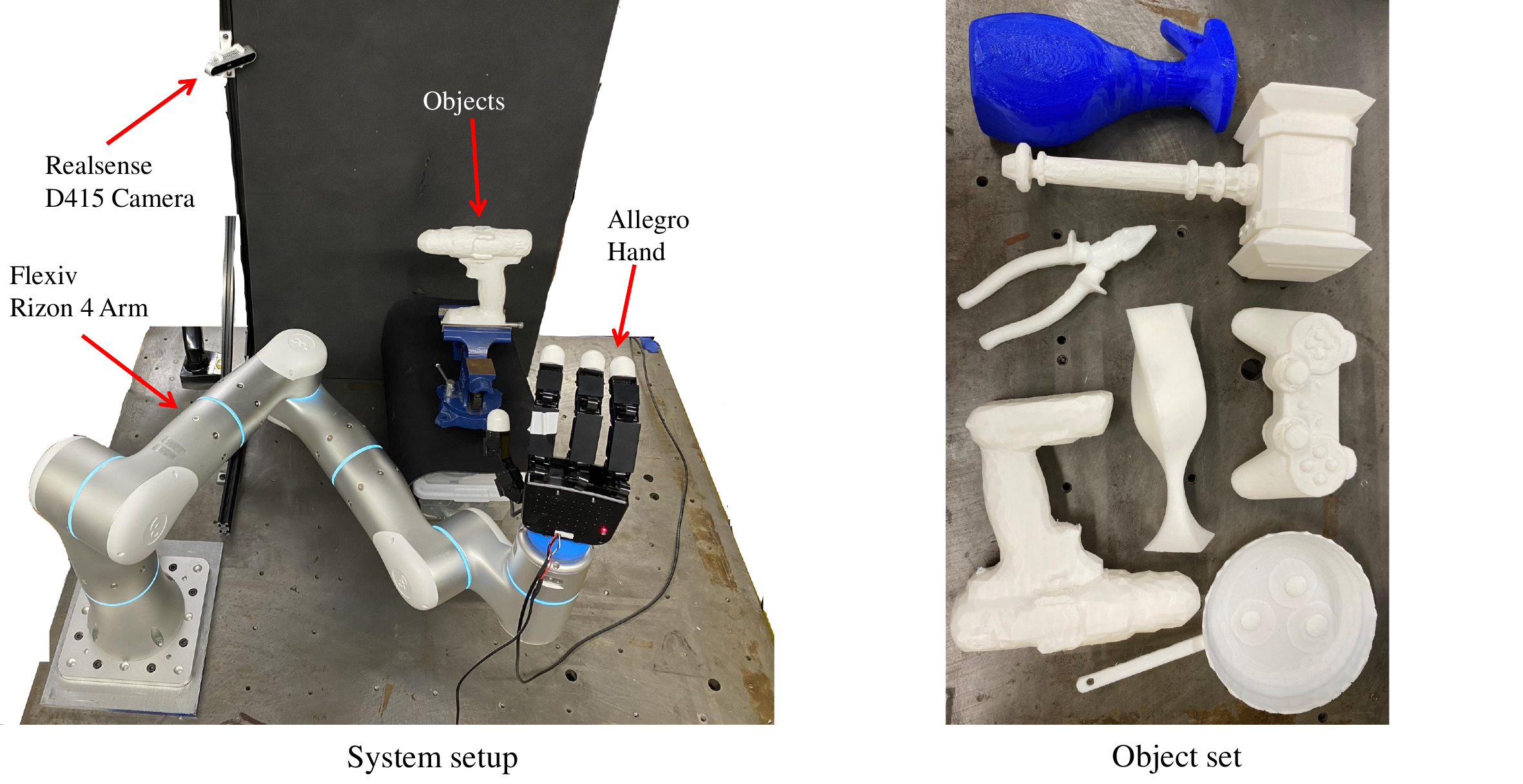}
\caption{
The illustration of our real world experiment settings.
}
\label{fig:real world seting}
\end{figure}

\begin{figure}[t]
\centering
\includegraphics[width=0.8\linewidth]{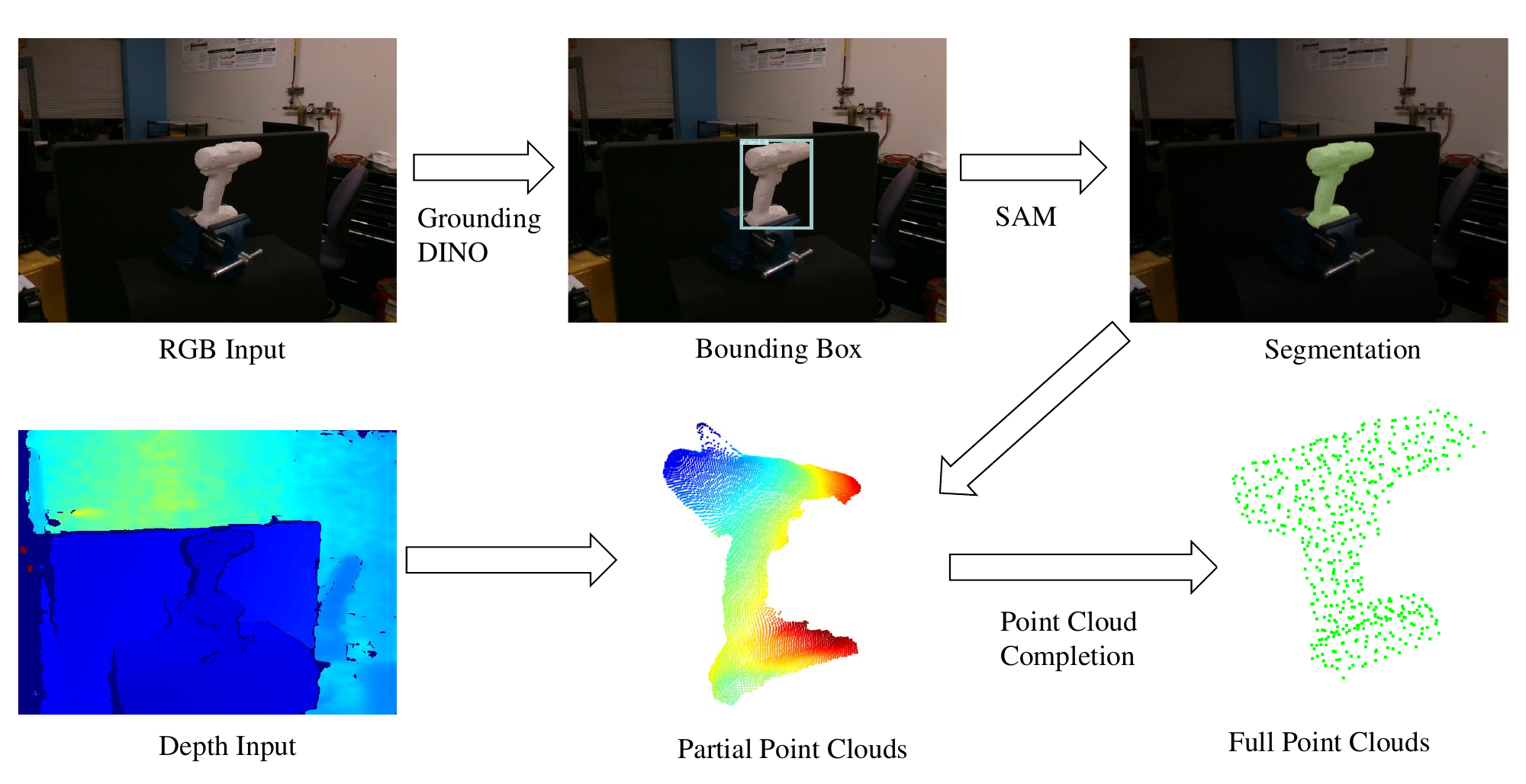}
\caption{
Real world experiment pipeline.
}
\label{fig: realword_pipline}
\end{figure}

\begin{figure}[t]
\centering
\includegraphics[width=\linewidth]{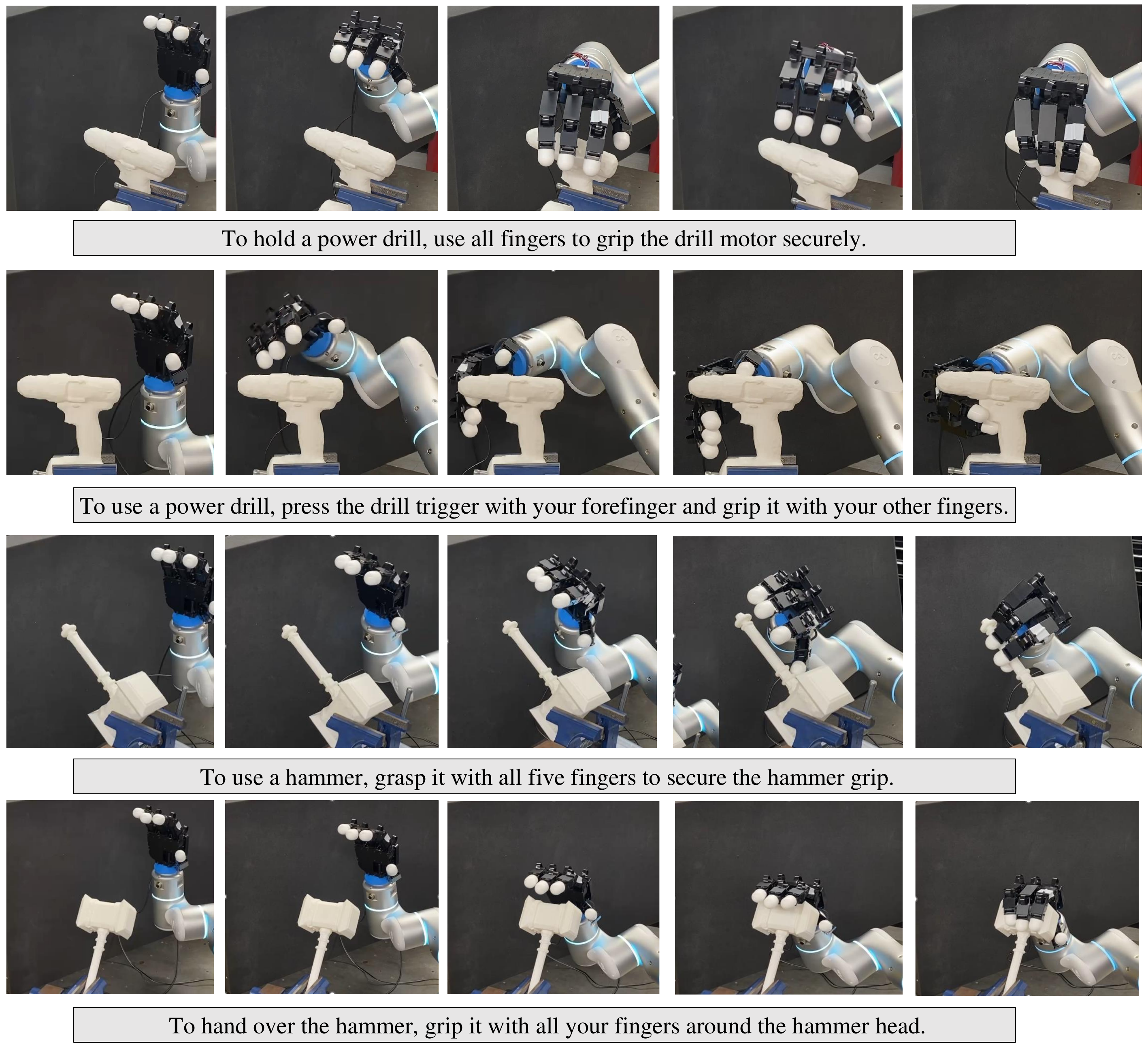}
\caption{
The visualization of real world experiments.
}
\label{fig:real world experiments}
\end{figure}

\subsection{Additional Experiments}
\subsubsection{Qualitative Experiments of Quality Grasp Component. }
We provide additional qualitative results to verify the effectiveness of Quality Grasp Component. Figure~\ref{fig: coarse2fine} shows the grasps before and after the application of the Quality Grasp Component, demonstrating that QGC can prevent object penetration and maintain consistency with the original intention.

\subsubsection{Qualitative Experiments of Task-oriented Guidance.}
We conduct qualitative experiments to demonstrate the generalization of our DexGYSGrasp framework to task-oriented or functional grasp task. Specifically, we input task-oriented guidance (e.g., "use" or "hold") into our framework, which has been trained on DexGYSNet. As shown in Figure~\ref{fig: simple_input}, our DexGYSGrasp framework exhibits good compatibility with these inputs. This further confirms that our approach enables more flexible and natural human-robot interactions.

\subsection{Real World Experiments Details}
\label{sec: realworld detials}
\textbf{Experimental Environment}
Figure \ref{fig:real world seting} shows the settings of our real world experiments. The experiments are conducted on Allegro hand, a Flexiv Rizon4 arm and an Intel Realsense D415 camera. The experimental object is a 3D printed object from test set of DexGYSNet.

\textbf{Experiment Pipeline}
Our DexGYSGrasp takes full point clouds as input following recent works in dexterous grasping~\cite{dexgraspnet, scene_diffuser, xu2024dgtr}. To make our methods more practical, we employ three off-the-shelf models in a cascade to obtain a full point cloud from scene point cloud. As shown in Figure~\ref{fig: realword_pipline}, we input the object category and the RGB image into an open-set detection model~\cite{liu2023groundingdino} to detect the bounding box of the object. This bounding box is then used as a prompt for SAM~\cite{kirillov2023SAM} to obtain the segmentation of the object. Next, we crop the target depth image using the segmentation map and the depth input. Finally, we convert the partial depth image into point clouds and feed it into a point completion network~\cite{yuan2018pcn} to obtain the final full point clouds. Then, the full point clouds are fed into our framework to obtain the dexterous grasp pose, which is then transformed into the real coordinate system. In execution, we first move the arm to the 6-DOF pose of the dexterous hand root node, and then control the joint angles to achieve the target pose.

\begin{table}[ht]
\centering
\begin{tabular}{@{}c*{7}{p{1.4cm}}@{}} 
\toprule
& Power drill & Hammer & Trigger sprayer & Game controller & Pincer & Frying pan & Wineglass  \\
\midrule
Success & 9/10 & 4/10 & 3/10 & 8/10 & 5/10 & 3/10 & 6/10  \\ 
\bottomrule
\end{tabular}
\vspace{0.5cm}
\caption{The results of real word experiment.}
\label{tab: realword}
\end{table}

\textbf{Experiment Results}
The experiment results are presented in Table \ref{tab: realword}. For each object, we command robot with different language instruction, and each instruction is tested five times, resulting in a total of ten grasping trials per object. A grasp is deemed successful if it aligns with the intended instruction and maintains stability, preventing the object from falling. Our method demonstrates a moderate success rate, indicating its effectiveness. Further research on real-world scenarios is recommended to enhance the robustness of our approach.

\subsection{Societal Impacts and Limitations}
The core innovation of this paper has a significant positive impact on society. We propose a novel task: language-guided dexterous grasp generation, which can promote human-robot interaction and expedite the deployment of robots in real-world scenarios. Additionally, we introduce an innovative framework to accomplish this task. Our approach can generate high-quality grasps while ensuring consistency of intent and diversity of grasps.

However, our method still faces some challenges in real-world deployment. Due to limitations in the current development of robotic arm control and physical structures, we cannot guarantee success in every grasp execution in the real world. In future work, we will further enhance the quality of grasp generation to improve the success rate in real-world scenarios.

\end{document}